\documentclass{article}
\usepackage{ijcai24}

\usepackage[dvipsnames]{xcolor}
\usepackage{times}
\usepackage{soul}
\usepackage{url}
\usepackage[utf8]{inputenc}
\usepackage[small]{caption}
\usepackage{graphicx}
\usepackage{amsmath}
\usepackage{amsthm}
\usepackage{booktabs}
\usepackage{algorithm}
\usepackage{algorithmic}
\usepackage{multirow}
\usepackage{array}
\usepackage[inkscapelatex=false]{svg}
\usepackage{fontawesome}

\title{KS-LLM: Knowledge Selection of Large Language Models with Evidence Document for Question Answering}

\author{
Xinxin Zheng$^{1,2}$
\and
Feihu Che$^3$\and
Jinyang Wu$^3$\and
Shuai Zhang$^3$\and
Shuai Nie$^2$\and
Kang Liu$^{1,2}$\and
Jianhua Tao$^{3,4}$\thanks{Corresponding author}\\
\affiliations
$^1$School of Artificial Intelligence, University of Chinese Academy of Sciences\\
$^2$Institute of Automation, Chinese Academy of Sciences\\
$^3$Department of Automation, Tsinghua University\\
$^4$Beijing National Research Center for Information Science and Technology, Tsinghua University\\
\emails
zhengxinxin2021@ia.ac.cn,
\{qkr, zhang\_shuai\}@mail.tsinghua.edu.cn,
wu-jy23@mails.tsinghua.edu.cn,
kliu@nlpr.ia.ac.cn,
nss90221@gmail.com,
jhtao@tsinghua.edu.cn
}
\begin{document}

\maketitle

\begin{abstract}
Large language models (LLMs) suffer from the hallucination problem and face significant challenges when applied to knowledge-intensive tasks. A promising approach is to leverage evidence documents as extra supporting knowledge, which can be obtained through retrieval or generation. However, existing methods directly leverage the entire contents of the evidence document, which may introduce noise information and impair the performance of large language models. To tackle this problem, we propose a novel Knowledge Selection of Large Language Models (KS-LLM) method, aiming to identify valuable information from evidence documents. The KS-LLM approach utilizes triples to effectively select knowledge snippets from evidence documents that are beneficial to answering questions. Specifically, we first generate triples based on the input question, then select the evidence sentences most similar to triples from the evidence document, and finally combine the evidence sentences and triples to assist large language models in generating answers. Experimental comparisons on several question answering datasets, such as TriviaQA, WebQ, and NQ, demonstrate that the proposed method surpasses the baselines and achieves the best results.
\end{abstract}

\section{Introduction}
Large language models (LLMs) have made significant progress in the field of natural language processing, achieving remarkable results in tasks such as text generation \cite{lin2023text}, machine translation \cite{moslem2023adaptive}, and dialogue systems \cite{sun2023generative}. However, despite notable successes in certain areas, LLMs suffer from severe hallucination problems, which may generate contents that deviate from the facts or contain fabricated information \cite{rawte2023survey}. It has always been a challenge for large language models to handle knowledge-intensive tasks \cite{petroni2021kilt}, such as question answering \cite{wu2022efficient,andrus2022enhanced} and fact checking \cite{atanasova2020generating}, since they may potentially provide incorrect or misleading information, leading to task failures or inaccurate results.

\begin{figure}[t]
  \centering
  \includegraphics[width=0.35\textwidth]{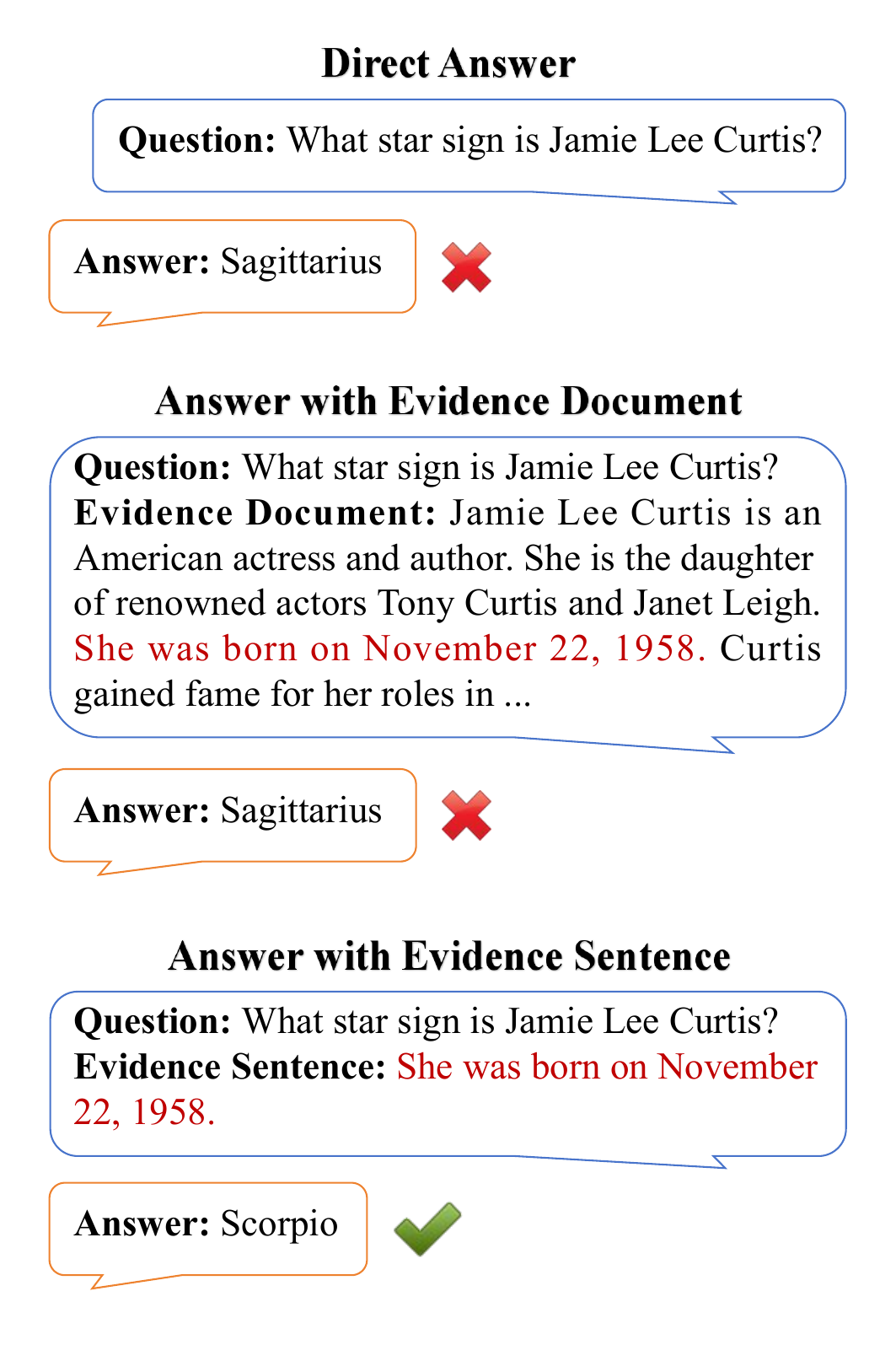}
  \caption{The large language model generates the incorrect answer with the given evidence document, while obtaining the correct answer with evidence sentences selected from the evidence document.}
  \label{fig1}
\end{figure}

In the question answering task, introducing supporting knowledge related to the input question can effectively alleviate the hallucination problem of large language models \cite{sun2023think}. Previous methods use evidence documents as the external supporting knowledge, providing extra information and validation for the model when generating answers. There are currently two main approaches for acquiring evidence documents: retrieval-based methods \cite{izacard2021leveraging,abdallah2023generator} and generation-based methods \cite{yu2023generate,sun2023recitationaugmented}. Retrieval-based methods involve retrieving evidence documents relevant to the input question from large-scale corpus, such as Wikipedia. In contrast, generation-based methods leverage the internal knowledge of large language models to generate evidence documents or background knowledge related to the input question. Existing results demonstrate that generation-based methods significantly improve the accuracy of answering questions, even without incorporating new external information. \cite{yu2023generate}.

Although the above methods provide extra knowledge for LLMs to better understand questions, they still suffer from two drawbacks. First, previous methods directly integrate all the contents in the evidence document into LLMs, which may lead to information overload and decrease the accuracy and efficiency of answering questions. Considering an evidence document that involves a large amount of contents, if LLMs need to process and understand the entire contents, they may struggle to accurately extract and utilize the knowledge relevant to the question. As shown in Figure 1, using the complete evidence document fails to facilitate large language models to answer the question correctly, while providing precise evidence sentences can lead to an accurate answer. Secondly, existing methods only use a single form of data source for knowledge augmentation \cite{gao2023retrieval}, ignoring the interaction and complementary relationship between different forms of knowledge. For example, structured knowledge can provide relations between entities, while textual knowledge can offer more detailed descriptions and contextual information.

Interestingly, when performing the question answering task, humans leverage their comprehensive capabilities to select key knowledge associated with the question from the evidence document to produce accurate answers. Inspired by this, we propose a novel Knowledge Selection of Large Language Models (KS-LLM) method, which aims to enhance the performance of large language models in the QA task by extracting relevant and useful knowledge from evidence documents. Specifically, we first construct triples based on the input questions, and then select evidence sentences from the evidence document that are most relevant to the triples. Finally, we incorporate the selected evidence sentences with constructed triples as supporting knowledge for LLMs to generate the final answer.

We conduct comprehensive experiments on three widely used datasets, i.e., TriviaQA-verified, WebQ, and NQ, using three representative large language models, i.e., Vicuna-13B, Llama 2-13B, and Llama 2-7B. Experimental results demonstrate that KS-LLM can significantly improve the performance of large language models on the question answering task, indicating that our method is capable of effectively selecting relevant knowledge from evidence documents for generating accurate answers.

In summary, our main contributions are as follows:
\begin{itemize}
\item We propose a novel method that can select knowledge snippets that are highly relevant to the input question from the evidence document, improving the accuracy and reliability of large language models in answering questions and alleviating the hallucination problem.
\item Our proposed method combines multiple forms of knowledge, including textual evidence sentences and structured triples, taking full advantages of the interaction and complementary relationship between different forms of knowledge.
\item We demonstrate the effectiveness of the proposed KS-LLM method in the QA task. Extensive experimental results show that our method surpasses different baselines and achieves the best performance on three datasets.
\end{itemize}

\section{Related Work}
\subsection{Question Answering with Evidence Documents}
Evidence documents typically refer to documents containing information relevant to the query question, which are used to facilitate accurate answers or support the reasoning process. Question answering methods with evidence documents are mainly divided into two categories: retrieval-based methods and generation-based methods.

Retrieval-based methods retrieve documents that may contain the answer strings from a large-scale corpus, and then use the retrieved documents to generate correct answers. Early research utilizes sparse retrieval methods, such as BM25 \cite{chen2017reading}, or neural ranking models \cite{guo2016deep,qaiser2018text} to retrieve documents. Representative works of early research include DrQA \cite{seo2016bidirectional} and BiDAF \cite{chen2017reading}. Subsequently, dense retrieval models like ORQA \cite{lee2019latent} and DPR \cite{karpukhin2020dense} are proposed, which encode contextual information to obtain dense representations of documents. Recent works \cite{qu2021rocketqa,raffel2020exploring} enhance the performance of retrievers to obtain more effective evidence documents, further improving the accuracy of models in answering questions. Rather than relying on external knowledge, generation-based methods extract knowledge from the parameters of large language models to generate evidence documents. Recent research shows that large-scale pre-trained models can form an implicit knowledge base after pre-training \cite{radford2018improving,yang2019xlnet}, which contains a vast amount of knowledge. GenRead \cite{yu2023generate} is the first work to propose using documents generated by large language models instead of retrieved documents. GenRead \cite{yu2023generate} and RECITE \cite{sun2023recitationaugmented} generate contextual documents with the implicit knowledge of large language models, and then read the documents to predict final answers.

Although evidence documents can provide additional knowledge to help answer questions, the above method utilizes all the information in the evidence documents as supporting knowledge, which may introduce noise irrelevant to the query question. Our proposed method effectively extracts the most relevant sentences from the evidence documents to assist the large language model, improving the accuracy and efficiency of answering questions.

\begin{figure*}[ht]
    \includegraphics[width=1\textwidth]{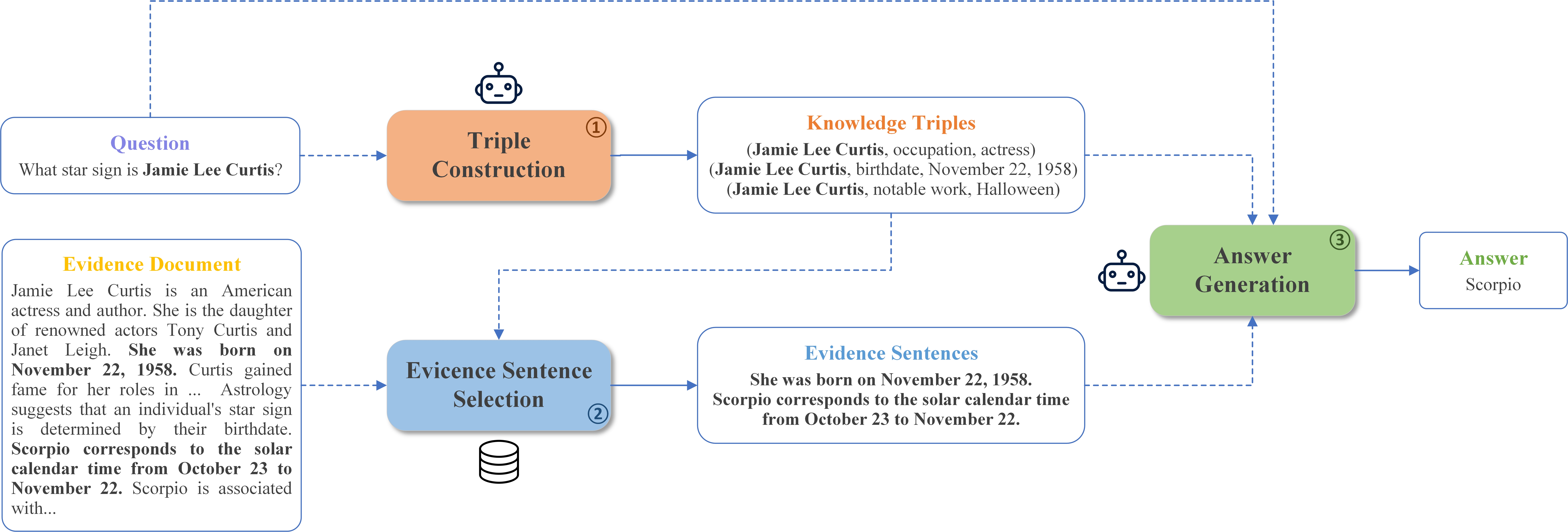}
    \caption{The proposed KS-LLM framework consists of three components: (1) triple construction, (2) evidence sentence selection, and (3) answer generation. The triple construction and answer generation steps are implemented by large language models, while the evidence sentence selection step is implemented by the vector database. The dashed line indicates the input of each step and the solid line indicates the output of each step. Given a question and its corresponding evidence document as input, our method can effectively extract valuable knowledge from the evidence document to acquire the correct answer.}
    \label{fig2}
\end{figure*}

\subsection{Question Answering with Knowledge Graphs}
Knowledge graphs (KGs) store factual knowledge in the real world, and have advantages in dynamic, explicit, and structured knowledge representation \cite{pan2023unifying}. Question answering methods with knowledge graphs utilize structured knowledge graphs as auxiliary information to improve the performance of question answering systems, usually involving knowledge bases such as Wikidata \cite{vrandevcic2014wikidata} and Freebase \cite{bollacker2008freebase}.

Early studies \cite{zhang2019ernie,peters2019knowledge,wang2021kepler} require models to learn structured knowledge in knowledge graphs during the training or fine-tuning process, which consumes a large amount of computing resources. Recent methods leverage knowledge by incorporating knowledge graphs into the prompts of large language models and express knowledge graphs in the form of triples. ToG \cite{sun2023think} explores entities and relations through external knowledge bases, dynamically discovering multiple reasoning paths on the knowledge graph to enhance the multi-hop reasoning capabilities of large language models. KGR \cite{guan2023mitigating} uses factual knowledge stored in the knowledge graph to correct errors that may occur during the reasoning process, which can automatically alleviate the hallucination problem of large language models. CoK \cite{li2023chain} leverages query languages to obtain knowledge from structured knowledge sources, improving the factual correctness of large language models.

Although the above methods improve the performance of large language models on the question answering task, they only utilize a single form of knowledge. Our proposed method simultaneously combines structured triples and textual sentences from evidence documents, taking full advantage of multiple forms of knowledge.

\section{Method}
The goal of this study is to enhance the performance of large language models on knowledge-intensive tasks by leveraging triples for effective knowledge selection from evidence documents. In this section, we present a detailed description of our proposed approach, KS-LLM, for solving QA tasks. As shown in Figure 2, KS-LLM consists of three stages: (1) triple construction, which generates a set of triples based on the subject entities in the query question; (2) evidence sentence selection, where the most relevant evidence sentences to the triples are extracted from the evidence document; (3) answer generation, which utilizes the triples and evidence sentences as supporting knowledge to generate the final answer. Next, we will describe each component in KS-LLM respectively.

\subsection{Triple Construction}
The process of triple construction employs the large language model to generate structured triples based on the natural language question, facilitating the precise capture of the intent and crucial information of the question. Given a query question $Q$, the process of triple construction aims to generate a set of triples $T=\{(h_i,\;r_i,\;t_i)\},\;i=1,...,m$ using the large language model, where $m$ is the number of triples, $h$ and $t$ are the head entity and tail entity respectively, and $r$ denotes the relation between the head entity and tail entity. Formally, $T$ is obtained by:
\begin{equation}
T = \mathbf{LM}(Q)
\end{equation}
where $\mathbf{LM}$ represents a specific large language model.

Taking a query question as input, the process of triple construction first identifies the subject entity in the query question, and then generates a set of triples with rich information based on the subject entity. Specifically, we extract the entity related to the topic of the query question, referred to as the subject entity. This entity can be individuals, locations, organizations, or other entities that reflect the core contents of the query question. Next, we construct a set of triples utilizing the subject entity as the head entity. The expanded triples cover various aspects of knowledge closely related to the query question, providing contextual information to the model from multiple perspectives. As illustrated in Figure 2, we first extract the subject entity \textit{Jamie Lee Curtis} from question \textit{``What star sign is Jamie Lee Curtis?''}, and then construct several triples with \textit{Jamie Lee Curtis} as the head entity, such as \textit{(Jamie Lee Curtis, occupation, actress)}.

By focusing on the subject entity, we ensure that the constructed triples capture the most relevant information necessary for answering the query question. The constructed triples not only help the model better comprehend the query question but also guide large language models in performing complex reasoning, ultimately generating accurate and consistent answers. The process of triple construction is automatically executed by large language models, without requiring additional manual efforts.

\subsection{Evidence Sentence Selection}
Evidence documents refer to documents that provide background information, relevant facts, or supporting knowledge to query questions. However, evidence documents typically contain a large amount of information, and inputting the entire document into a large language model may introduce noise information, making it more difficult for the model to understand and filter relevant knowledge. Therefore, it is crucial to select valuable evidence sentences from evidence documents, which can significantly improve the quality and accuracy of large language models on the question answering task. 

Given the constructed triples $T$ and an evidence document $ D = (s_1, s_2, ..., s_n) $, where $s$ represents a sentence and $n$ is the number of sentences, the process of evidence sentence selection extracts the evidence sentences $S$ most relevant to the triples $T$ from the evidence document $D$. Specifically, we initially employ the BERT \cite{kenton2019bert} model to obtain the embedding representations of constructed triples and each sentence in the evidence document. This can be formulated as:
\begin{equation}
q = \mathbf{Bert}(T), K = \{k_i | k_i = \mathbf{Bert}(s_i)\}
\end{equation}
where $q$ and $K$ denote the embeddings of triples and the evidence document respectively. The BERT model captures the semantic information and contextual features of sentences by encoding them into dense vectors. Subsequently, in order to measure the semantic similarity, we calculate the Euclidean distance between each sentence and the triples based on embedding representations, and select top $k$ sentences with the closest distance as evidence sentences. The indices of evidence sentences are calculated by:
\begin{equation}
L = \arg\min_i^k \{\|k_i-q\|_2\}
\end{equation}
Here, $L=(l_1, l_2, ..., l_k)$ represents the indices of the top $k$ minimum values returned by $\overset k{\arg\min}$, and $\left\|\cdot\right\|$ denotes the Euclidean distance. Finally, $S = \{s_{l_i} | l_i\in L\}$ is the evidence sentences selected from the evidence document. We set $k=2$ in the experiments.

As shown in Figure 2, we compute the Euclidean distance between the triples \textit{(Jamie Lee Curtis, occupation, actress), (Jamie Lee Curtis, birthdate, November 22, 1958), (Jamie Lee Curtis, notable work, Halloween)} and each sentence in the evidence document. Then we select the top two sentences with the closest distances as the evidence sentences, i.e., \textit{``She was born on November 22, 1958.''} and \textit{``Scorpio corresponds to the solar calendar time from October 23 to November 22.''}.

During the evidence sentence selection step, we can extract the most relevant evidence sentences to the triples from voluminous evidence documents. These sentences contain crucial information related to the query question and provide supporting knowledge for subsequent answer generation. In addition, compared to directly using the entire document as evidence, effective evidence sentence selection eliminates irrelevant information in the evidence document that may hinder the answer. The evidence sentence selection process is implemented through the vector database, which offers the advantage of high efficiency.

\subsection{Answer Generation}
We integrate the triples and evidence sentences as supporting knowledge, combine them with the query question, and leverage the reasoning capability of large language models to obtain the final answer. Formally, the final answer $A$ is generated by:
\begin{equation}
A = \mathbf{LM}(Q, T, S)
\end{equation}

Previous methods \cite{sun2023recitationaugmented,sun2023think} only utilize knowledge graphs or evidence documents as external knowledge to assist large language models in question answering, without considering the interaction between different forms of knowledge. The triples provide structured knowledge in knowledge graphs, while evidence sentences provide detailed information from the evidence document in textual format. By fusing multiple forms of knowledge at different granularities, we are able to provide the model with richer context and factual knowledge, facilitating large language models to generate more accurate and consistent answers.

Because different models with different sizes possess varying abilities to follow instructions, the output format control in the prompt may differ slightly when generating answers. We expect the model to generate a single entity as the answer so that a fair comparison can be made.

\begin{table*}[htbp]
    \resizebox{\linewidth}{!}{
    \begin{tabular}{p{2.4cm}<{\centering}p{1.3cm}<{\centering}p{1.3cm}<{\centering}p{1.3cm}<{\centering}p{1.3cm}<{\centering}p{1.3cm}<{\centering}p{1.3cm}<{\centering}p{1.3cm}<{\centering}p{1.3cm}<{\centering}p{1.3cm}<{\centering}}
    \toprule
     & \multicolumn{3}{c}{TriviaQA-verified} & \multicolumn{3}{c}{WebQ} & \multicolumn{3}{c}{NQ} \\
     \cmidrule(r){2-4} \cmidrule(r){5-7} \cmidrule(r){8-10}
     & Vicuna -13B & Llama 2 -13B & Llama 2 -7B & Vicuna -13B & Llama 2 -13B & Llama 2 -7B & Vicuna -13B & Llama 2 -13B & Llama 2 -7B \\
    \midrule
    \multicolumn{10}{c}{\textit{Without evidence document}} \\
    \midrule
    Standard & 51.45 & 62.07 & \textbf{51.03} & 17.91 & 23.18 & 17.08 & 14.93 & 16.59 & 15.10 \\
    \midrule
    \multicolumn{10}{c}{\textit{With evidence document}} \\
    \midrule
    Standard+doc & 52.69 & 56.97 & 49.24 & 18.31 & 23.47 & 17.52 & 17.04 & 21.06 & 19.39 \\
    CoT+doc & 50.34 & 55.45 & 49.52 & 18.26 & 22.98 & 18.46 & 17.12 & 20.16 & 20.36 \\
    KS-Q & 52.10 & \textbf{62.76} & 49.66 & 20.47 & 24.66 & 19.29 & 15.07 & 20.00 & 17.92 \\
    KS-T & 52.28 & 53.66 & 40.59 & 21.79 & 23.82 & 20.42 & 15.40 & 16.92 & 11.25 \\
    KS-S & 51.59 & 57.79 & 44.55 & 20.23 & 24.11 & 18.85 & 15.62 & 18.89 & 17.26 \\
    KS-LLM (Ours) & \textbf{58.48} & 55.72 & 44.14 & \textbf{21.85} & \textbf{24.70} & \textbf{21.11} & \textbf{17.59} & \textbf{21.69} & \textbf{20.95} \\
    \bottomrule
    \end{tabular}}
    \caption{Performance comparison on TriviaQA-verified, WebQuestions (WebQ), and Natural Questions (NQ) datasets using three large language models with different sizes. We report the exact match (EM) score in the table, and the best result for each model is \textbf{bolded}.}
    \label{tab1}
\end{table*}

\section{Experiments}
In this section, we conduct comprehensive experiments of our proposed KS-LLM method on the QA task with evidence documents. We report empirical evaluations of KS-LLM on three widely adopted datasets: TriviaQA-verified \cite{joshi2017triviaqa}, WebQ \cite{berant2013semantic}, and NQ \cite{kwiatkowski2019natural}. Following previous works, we use the exact match (EM) score to evaluate the model performance on the QA task. We also evaluate the effectiveness of KS-LLM on two different base LLMs with various sizes: Vicuna \cite{zheng2023judging} and Llama 2 \cite{touvron2023llama}.

\subsection{Experimental Setup}
\subsubsection{Datasets and Evaluation.}
We conduct experiments on three representative QA datasets.

\textbf{TriviaQA-verified} \cite{joshi2017triviaqa} is a reading comprehension dataset that covers a wide range of topics, consisting of over 650K question-answer-evidence triples. The evidence documents are collected by remote supervision and may include noise which is irrelevant to the question. Therefore, in this paper, we utilize the verified set in TriviaQA, which is manually verified to ensure that each document contains the relevant facts necessary for answering the question.

\textbf{WebQ} \cite{berant2013semantic} refers to WebQuestions, which is an open-domain question answering dataset containing numerous question-answer pairs. WebQ includes questions that are sourced from the web, aiming to evaluate the performance of QA systems in handling real-world questions without domain restrictions.

\textbf{NQ} \cite{kwiatkowski2019natural} refers to Natural Questions, which is a widely used open-domain question answering dataset created by the Google AI team. This dataset contains real-world questions selected from Google search logs and is of significant importance for evaluating and advancing research in question answering systems.

Due to the absence of evidence documents in WebQ and NQ datasets, we follow the previous work \cite{yu2023generate} and employ the large language model to generate an evidence document for each question in WebQ and NQ. Specifically, we use Vicuna 13B for evidence document generation.

We report the exact match (EM) score respectively on the validation set of each dataset in order to evaluate the model performance. An answer is considered correct if and only if its normalized form has a match in the acceptable answer list.

\subsubsection{Fundamental Models.} 
We conduct experiments on three representative fundamental large language models with various sizes.

\textbf{Vicuna} \cite{zheng2023judging} is an open-source large language model launched by the Large Model Systems Organization (LMSYS Org). Vicuna includes three versions: 7B, 13B, and 33B, and is fine-tuned based on Llama with the open conversation dataset collected by SharedGPT. The latest release, Vicuna 1.5, is fine-tuned based on Llama 2 and supports inputs with a maximum context length of 16K. We utilize the 13B versions of Vicuna 1.5.

\textbf{Llama 2} \cite{touvron2023llama} is an open-source large language model released by Meta Company, including three versions: 7B, 13B, and 70B. Llama 2 is trained on datasets comprising over 2 trillion tokens, and the fine-tuning data includes publicly available instruction datasets, along with over 1 million new annotated examples. We utilize the 7B and 13B versions of Llama 2.

\subsubsection{Baselines.}
We set up six different baselines.

\textbf{Standard} baseline prompts the large language model to directly output answers to the questions.

\textbf{Standard+doc} baseline combines the question and the corresponding evidence document as input, prompting the large language model to output the answer. For the Trivia-verified dataset, since the length of its evidence documents may exceed the maximum input length of the model, we set a unified parameter $max\_token$ to limit the length of evidence documents. In this paper, we set $max\_token$ to 300.

\textbf{CoT+doc} baseline follows the same setup as Standard+doc baseline while incorporating the Chain of Thought (CoT) \cite{wei2022chain} approach.

\textbf{KS-Q} calculates the embeddings of the question and each sentence from the evidence document, and selects the top $k$ sentences most similar to the question as evidence sentences. Subsequently, KS-Q inputs the question and evidence sentences into the large language model. Instead of using the question, our approach leveraging triples to select evidence sentences.

\textbf{KS-T \& KS-S} baselines utilize the triples generated in the triple construction step and the evidence sentences obtained in the evidence sentence selection step as their supporting knowledge, respectively. In contrast, our proposed method integrates both triples and evidence sentences as supporting knowledge. Then KS-T and KS-S respectively input questions and triples, questions and evidence sentences into the large language model.

\subsection{Main Results}
As reported in Table 1, the proposed KS-LLM method demonstrates superior performance by outperforming multiple baselines and achieving significant advancements across all three datasets. Specifically, in the case of utilizing open-source models, the KS-LLM method achieves impressive EM scores of 58.48, 24.7, and 21.69 on the Trivia-verified, WebQ, and NQ datasets, respectively. Moreover, the KS-LLM method still maintains superior performance compared to methods with evidence documents. For example, compared to the Cot+doc method using Vicuna-13B, our KS-LLM yields substantial enhancements of 8.14 and 3.59 on the Trivia-verified and WebQ datasets, respectively. These results fully demonstrate that our method effectively extracts valuable knowledge from evidence documents, thereby significantly improving the accuracy of large language models in answer generation. Furthermore, our method outperforms the KS-T and KS-S methods, which solely exploit a single form of knowledge, in the vast majority of cases. This indicates that integrating different forms of knowledge enables the effective utilization of the interaction and complementary relationship between knowledge, further enhancing the knowledge absorption capability of large language models.

We also discover that directly incorporating appropriate evidence documents often leads to minor performance improvements (Standard v.s. Standard+doc). In seven out of nine cases across three datasets using three large language models, incorporating evidence documents results in higher accuracy for the large language models. However, applying the chaining of thought (CoT) technique does not consistently enhance the performance of large language models (Standard+doc v.s. CoT+doc). This could be due to the use of 0-shot prompt in our experiments, where the performance of 0-shot CoT is not stable. Moreover, the choice of fundamental models significantly affects the utilization of knowledge. For example, the Llama 2 model struggles to effectively utilize valid knowledge on the TriviaQA-verified dataset. This may be attributed to the fact that evidence documents for TriviaQA-verified are obtained from the web through distant supervision \cite{joshi2017triviaqa} and contain non-typical natural language expressions, such as ``Sam Smith releases new James Bond title song | Film | DW.COM | 25.09.2015'', which Llama 2 is not adapted to handle such form of knowledge.

\begin{table}[htbp]
    \centering
    \begin{tabular}{cccc}
        \toprule
        Method & Length (token) & EM & Time \\
        \midrule
        Standard & 0 & 51.45 & 3min40s \\
        Standard+doc & 300 & \underline{52.69} & 4min41s \\
        Standard+doc & 500 & 49.52 & 5min22s \\
        Standard+doc & 1000 & 47.45 & 7min10s \\
        Standard+doc & 2000 & 37.66 & 11min31s \\
        KS-LLM & - & \textbf{58.48} & - \\
        \bottomrule
    \end{tabular}
    \caption{Impact of evidence document length on TriviaQA-verified dataset with Vicuna-13B. Time refers to the running time of the large language model for inference on NVIDIA A100 80G. The best result in baselines is \underline{underlined} and the best result overall is \textbf{bolded}.}
    \label{tab2}
\end{table}

\subsection{Impact of Evidence Document Length}
The length of external knowledge may affect the performance of large language models. Providing appropriate external knowledge can enhance the performance of the model, while excessively long knowledge may degrade the performance of large language models. The length of evidence document can vary significantly in the real world, making it necessary to select the appropriate document length to facilitate large language models in answering questions.

To investigate the impact of different evidence document lengths on large language models, we conduct experiments using Vicuna-13B on the TriviaQA-verified dataset. Specifically, we report the performance of the Standard+doc baseline under different evidence document lengths and compare them with the Standard baseline and the proposed KS-LLM method. In addition, we also report the running time required for inference with various lengths of documents on NVIDIA A100 80G. Through this experiment, we aim to gain a deeper understanding of the adaptability of large language models under different evidence document lengths. This is of great help in choosing the optimal evidence document length in practical applications, balancing the requirements of performance and efficiency.

As shown in Table 2, the performance of large language models in answering questions is significantly influenced by the length of evidence document. Using a 300-token-length evidence document resulted in a 1.24 increase in the evaluation metric compared to not using any evidence document. However, as the length of the evidence document increases, there is a corresponding decrease in performance. This is consistent with the hypothesis from previous research that appropriate external knowledge can improve the performance of the model, while excessively long knowledge has a negative impact on the performance of large language models. The length of the evidence document also affects the inference time of large language models. As the length of the evidence document increases, the inference time proportionally extends. Using a 2000-token-length evidence document takes approximately three times longer for inference compared to not using any evidence document.

\begin{figure}[htbp]
  \centering
  \includegraphics[width=0.45\textwidth]{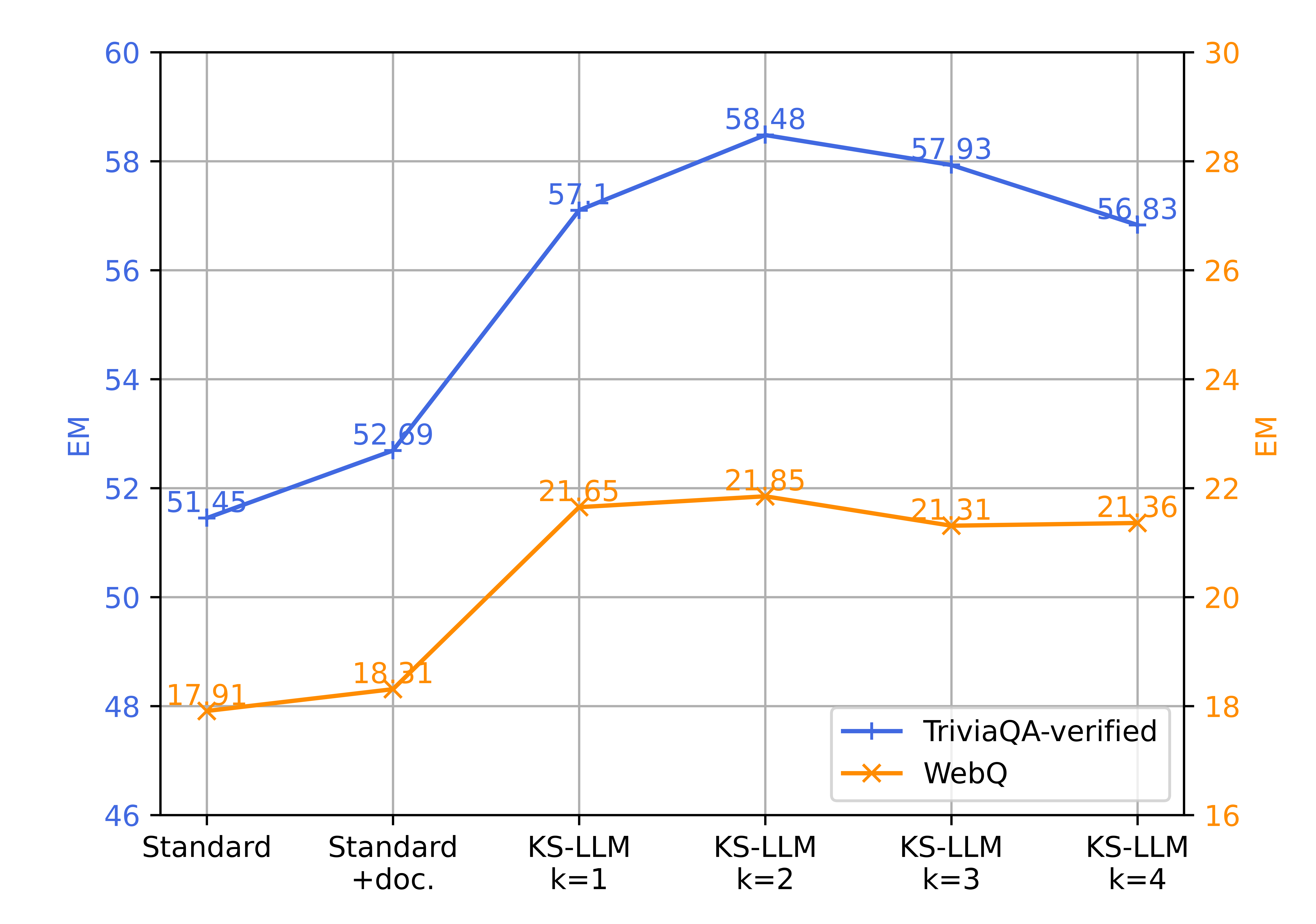}
  \caption{Impact of parameter $k$ on TriviaQA-verified and WebQ datasets with Vicuna-13B. We report the exact match (EM) score in the table.}
  \label{fig3}
\end{figure}

\begin{table*}[ht]
    \resizebox{\linewidth}{!}{
    \begin{tabular}{cccc}
    \toprule
    \multicolumn{4}{l}{\textbf{Question:} For which team did Babe Ruth blast his last Major League home run?} \\
    \multicolumn{4}{l}{\textbf{Answer:} Boston Braves} \\
    \multicolumn{4}{l}{\begin{tabular}[c]{@{}l@{}}\textbf{Output:} Babe Ruth (Standard) \faTimes; Philadelphia Athletics (Standard+doc) \faTimes;\\ \hspace{35pt} Philadelphia Athletics (CoT+doc) \faTimes; Boston Braves (KS-LLM) \faCheck \end{tabular}} \\
    \midrule
     & Triple Construction & Evidence Sentence Selection & Answer Generation \\
    \midrule
    \multicolumn{1}{m{0.1\textwidth}<{\centering}}{Intermediate Output of KS-LLM} & \multicolumn{1}{m{0.35\textwidth}}{(Babe Ruth, played for, Boston Red Sox), (Babe Ruth, played for, New York Yankees), (Babe Ruth, played for, Baltimore Orioles), (Babe Ruth, played for, St. Louis Browns), (Babe Ruth, played for, \textcolor{BrickRed}{Boston Braves})} & \multicolumn{1}{m{0.35\textwidth}}{On May 25, 1935, with the team on a road trip and playing at Forbes File in Pittsburgh, Ruth hammered three home runs and a single, driving in six runs. In 1935, Babe Ruth was forty years old, in poor physical shape, and playing out the string with the \textcolor{BrickRed}{Boston Braves}.} & \textcolor{BrickRed}{Boston Braves} \\
    \bottomrule
    \end{tabular}}
    \caption{Case study on the TriviaQA-verified dataset with Vicuna-13B. Answer strings that appear during the inference process of the KS-LLM approach are marked in \textcolor{BrickRed}{red}.}
    \label{tab3}
\end{table*}

\subsection{Impact of Parameter $k$}
The parameter $k$ represents the number of sentences selected from the evidence document during evidence sentence selection process. We extract the top $k$ sentences with the highest semantic similarity score to the triples from the evidence document and use them for the subsequent answer generation process. The parameter $k$ indicates how the quantity of supporting knowledge affects the performance of large language models. If $k$ is too small, large language models may not have sufficient knowledge for reasoning. If $k$ is too large, additional noisy knowledge may be introduced, interfering with the decision-making of large language models. 

We evaluate the impact of parameter $k$ on the performance of large language models using Vicuna-13B on Trivia-verified and WebQ datasets. Specifically, we report the performance of our KS-LLM method under different $k$ values while comparing with the Standard baseline and the Standard+doc baseline. Through this experiment, we were able to determine the optimal value of $k$, balancing the quantity of supporting knowledge and the risk of introducing noisy knowledge, which is conducive to improving the performance of large language models.

From Figure 3, it can be observed that the KS-LLM method achieves the best performance on both datasets when k=2, reaching 58.48 and 21.85, respectively. As the value of parameter k increases, there is a gradual decline in the performance of the model. This indicates that the parameter k plays an important role in the process of evidence sentence selection, and the number of evidence sentences directly affects the accuracy of large language models. When the parameter k is set to 2, we are able to achieve the best results using large language models in the question answering task. Furthermore, across different values of k, the proposed KS-LLM method consistently outperforms the Standard baseline and Standard+doc baseline. In the case of utilizing evidence documents, compared with the Standard+doc baseline, the KS-LLM approach achieves a maximum performance improvement up to 5.79 on the TriviaQA-verified dataset, while in the WebQ dataset, the maximum improvement reached 3.94. This demonstrates that effective knowledge selection from evidence documents can significantly enhance the performance of large language models, showcasing the superiority of our KS-LLM method.

\subsection{Case Study}
To better understand how the proposed KS-LLM method works, we provide a detailed example in Table 3. Given a question \textit{``For which team did Babe Ruth blast his last Major League home run?''} as input, the large language model gets the incorrect answer \textit{Babe Ruth} by directly answering the question. Given the question and its corresponding evidence document as input, the large language model yields the wrong answer \textit{Philadelphia Athletics} even if the evidence document contains the correct answer string \textit{Boston Braves}. As for the KS-LLM method, it first indicates that Babe Ruth played for multiple teams in the triple construction step, such as \textit{(Babe Ruth, played for, Boston Braves)}. Then, in the evidence sentence selection step, the crucial sentence \textit{``In 1935, Babe Ruth was forty years old, in poor physical shape, and playing out the string with the Boston Braves.''} containing the answer string is successfully identified from the evidence document. Finally, our proposed KS-LLM approach generates the precise answer \textit{Boston Braves} according to the triples and evidence sentences.

Through this example, it can be fully demonstrated that large language models may not be able to effectively utilize the contents of evidence documents, while KS-LLM can extract valuable information from the evidence documents to further generate accurate answers.

\section{Conclusion}
In this paper, we introduce KS-LLM, a novel knowledge selection method for large language models designed to tackle the question answering problem. Given the corresponding evidence documents, the KS-LLM approach effectively identifies the knowledge relevant to the question from evidence documents, thereby enhancing the performance and efficiency of large language models in the question answering task. The proposed method first constructs triples according to the query question, then extracts sentences from the evidence document that are most similar to the triples as evidence sentences, and finally integrates the triples and evidence sentences into the input of large language models to generate accurate answers. Experimental results demonstrate that our method achieves remarkable improvements on three datasets, indicating that KS-LLM is capable of selecting valuable knowledge snippets from evidence documents to assist large language models in answering questions.

\bibliographystyle{named}
\bibliography{ijcai24}

\end{document}